\documentclass[journal]{IEEEtran}
\pdfoutput=1
\usepackage{amsmath,amsfonts}
\usepackage{algorithmic}
\usepackage{algorithm}
\usepackage{array}
\usepackage[caption=false,font=normalsize,labelfont=sf,textfont=sf]{subfig}
\usepackage{textcomp}
\usepackage{stfloats}
\usepackage{url}
\usepackage{verbatim}
\usepackage{graphicx}
\usepackage{cite}
\usepackage{caption}
\newtheorem{lemma}{Lemma}
\newtheorem{theorem}{Theorem}

\usepackage{amssymb}
\usepackage{amsmath}
\usepackage{pythonhighlight}

\begin{document}

\title{Instance-aware Model Ensemble With Distillation For Unsupervised Domain Adaptation}

\author{Weimin Wu, Jiayuan Fan, Tao Chen, Hancheng Ye, Bo Zhang, Baopu Li
}


\maketitle

\begin{abstract}
The linear ensemble-based strategy (\textit{i.e.}, averaging ensemble) has been proposed to improve the performance in unsupervised domain adaptation (UDA) task. However, a typical UDA task is usually challenged by dynamically changing factors, such as variable weather, views and background in the unlabeled target domain. Most previous ensemble strategies ignore UDA's dynamic and uncontrollable challenge, facing limited feature representations and performance bottlenecks. To enhance the model adaptability between domains and reduce the computational cost when deploying the ensemble model, we propose a novel framework, namely Instance-aware Model Ensemble With Distillation (IMED), which fuses multiple UDA component models adaptively according to different instances and distills these components into a small model. The core idea of IMED is a dynamic instance-aware ensemble strategy, where for each instance, 
a non-linear fusion sub-network is learned 
that fuses the extracted features and predicted labels of multiple component models. The non-linear fusion method can help the ensemble model handle dynamically changing factors.
After learning a large-capacity ensemble model with good adaptability to different changing factors, we leverage the ensemble teacher model to guide the learning of a compact student model by knowledge distillation. Furthermore, we provide the theoretical analysis on the validity of IMED for UDA. Extensive experiments conducted on various UDA benchmark datasets (\textit{e.g.,} Office-31, Office-Home, and VisDA-2017) show the superiority of the model based on IMED to the state-of-the-art methods under the comparable computation cost.
\end{abstract}


\section{Introduction}
\begin{figure*}[t]
\centering
\includegraphics[width=180mm]{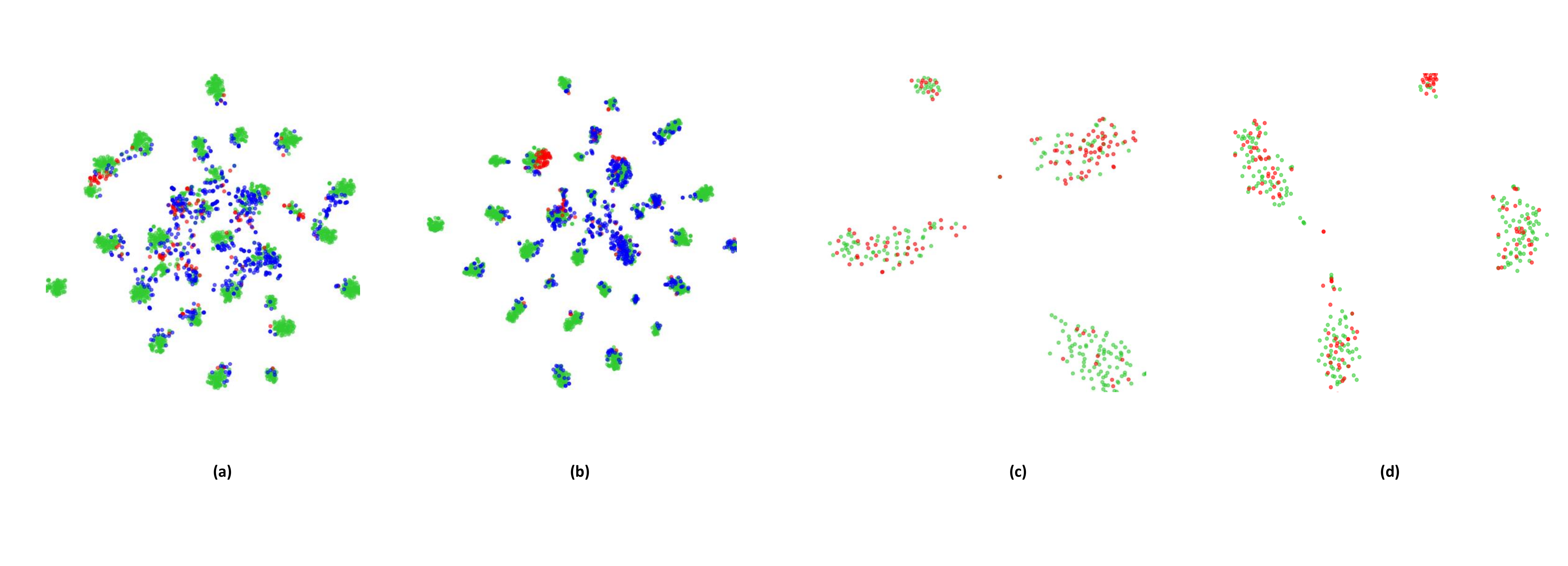}
\caption{(a) and (b) demonstrate the results of JAN and CDANs with random seed 5 from Webcam to Amazon of the dataset Office-31. In (a), red points are the instances that JAN correctly classifies while CDANs not, green and blue points present the correctly and wrongly classified instances by JAN. In (b), red points are the instances that CDANs correctly classifies while JAN not, green and blue points present the correctly and wrongly classified instances by CDANs. (c) and (d) use model from averaging ensemble and our proposed ensemble method respectively, based on the instances that only one of two models (JAN and CDANs) can classify correctly. Green and red points denote the instances that the model classifies correctly and wrongly. The percentage of green points in (c) and (d) is $63 \%$ and $64 \%$ respectively.}
\label{vis_class}
\end{figure*}

\IEEEPARstart{U}{nsupervised} domain adaptation (UDA) aims to transfer a deep network from a source domain to a target domain, where only unlabeled data is available in the target domain. Recent UDA has advanced a lot due to the success of deep learning, and can be mainly categorized into two classes. One calculates the distribution discrepancy between two domains and uses the final loss function to decrease it, such as joint adaptation network (JAN)~\cite{4}. The other applies adversarial learning technique to domain adaptation in a two-player game similarly to generative adversarial networks (GANs)~\cite{5}, such as conditional domain adversarial networks (CDANs)~\cite{7}. However, in the UDA field, there is no agreement on which model is the best. One learner could perform better than others in part of the target domain (\textit{i.e}. images with certain instances or background), while a further method could outperform others in another part (\textit{i.e}. images with other instances or background), as validated by Fig. \ref{vis_class}. For this purpose, ensemble learning could help different models to be complementary, which provides more accurate and robust performance for UDA.

Ensemble learning ~\cite{46, 47} as an effective machine learning method, has shown its superiority in improving the performance of various deep learning-based vision tasks. Traditionally, averaging the output of various component models is a valid ensemble way, while it faces the challenge of the large number of parameters and high computation cost. Therefore, knowledge distillation (KD) ~\cite{9, 44, 45}, which tries to train a student model to mimic the teacher's output, can help distill the ensemble model into a small student model. The student model not only achieves ideal results but also decreases computation cost. Recently, only a few works~\cite{12, 13} attempted to introduce ensemble model with distillation to UDA. These works demonstrate how an ensemble UDA model can be trained to learn complementary information and meanwhile transfer knowledge to a compact student model. However, these existing ensemble distillation models may still suffer from two bottlenecks. 

Firstly, most existing ensemble distillation models in UDA ~\cite{47, 48, 49, 50, 38, 39} simply use the averaging ensemble method, which means that the feature fusion treats all the instances equally important and does not differentiate them, ignoring that each instance has its own representation characteristics and present different capability for learning the ensemble model. Thus the feature fusion should be able to dynamically adjust its parameters conditioned on the instance. To solve this problem, we design a fully connected network, named endogeny network, to predict the instance-aware parameters of a fusion sub-network, whose input is the features and labels from all the component models. The fusion sub-network achieves dynamic adaptation according to each instance by the learned fusion parameters.

Secondly, the UDA has variable and unpredictable target domains and the space spanned by the object features from the target domain is very large. As a result, our designed feature fusion from the original component models should consider this to produce a representative but flexible-to-adapt feature for each instance, and map the target domain into a diversified and large enough feature space. This can enhance the ensemble model's representative capability for uncontrollable target domain and facilitate its transfer between domains. However, existing ensemble distillation methods ~\cite{47, 48, 49, 50, 38, 39} mainly adopt static fusion, which means a group of fixed fusion coefficients are pre-assigned to the component models and yields a limited feature space. For example, fixed weight is assigned to the pseudo labels from component models when fusing them ~\cite{47}. On the contrary, our designed feature fusion sub-network mentioned before adopts dynamic nonlinear integration of the features, and also develops a shuffle linear strategy to produce more diversified and representative features.


In view of the above, we propose a general framework of Instance-aware Model Ensemble With Distillation (IMED) which includes an ensemble model, domain adversarial discriminator and knowledge distillation module to solve the UDA problem. In particular, in the ensemble model, we tackle the two aforementioned challenges by designing an instance-aware nonlinear ensemble module. A fully connected network, named endogeny network, is used to predict the parameters of a fusion sub-network conditioned on features and labels from every component model. The sub-network can achieve non-linear fusion of features by the designed linear layer and non-linear layer (\textit{i.e.}, ReLU layer). Note that as there are so many parameters in fully connected linear layer that it is hard to predict them based on features and labels from component models alone. Thus we design a shuffle linear layer to replace fully connected linear layer in sub-network to reduce the parameters. The main idea of shuffle linear is to divide the input and output feature vectors into groups, and then designing fully connected linear layer within each group and using shared parameters among different groups, to produce diversified but representative features.



We validate our designed method based on six kinds of ensemble models, including the ensemble of two state-of-the-art (SOTA) CDANs with minimum class confusion and smooth domain adversarial training (CDAN+MCC+SDAT) ~\cite{15} using different random seeds. Other five models are the ensemble of JAN and CDANs, the ensemble of two CDANs with different random seeds, the ensemble of three CDANs with different random seeds, the ensemble of two ResNet-50 with different random seeds, and the ensemble of two ViT-B/16 with different random seeds. We evaluate the proposed framework on three public UDA datasets: Office-31, Office-Home, and VisDA-2017, to demonstrate the effectiveness of our framework. The contributions are summarized as follows:
\begin{itemize}
\item We explore the UDA task from a novel perspective of adaptive non-linear ensemble distillation conditioned on each instance, and correspondingly develop an IMED framework to achieve both robust and effective UDA. We also provide a theoretical proof to better understand why the proposed framework performs well.
\item We propose a non-linear instance-aware feature ensemble model for different component models, which can handle the dynamically changing factors among instances, span the feature representation space, and decrease computational cost. We also develop a shuffle linear module to greatly reduce the ensemble computation cost while maintaining good performance.
\item Experiments show that the proposed ensemble distillation model based on CDANs with minimum class confusion and smooth domain adversarial training (CDAN+MCC+SDAT), exceeds the state-of-the-art models on all three datasets. We also evaluate our framework under various configurations, and demonstrate that it is more effective than single component model under comparable computation cost.
\end{itemize}

\section{Related Work}
\subsection{Unsupervised Domain Adaptation} 
Domain adaptation ~\cite{51, 52, 16} aims to generalize a model across different domains with different distributions, it is widely applied in computer vision ~\cite{47, 18}. Main methods to decrease domain discrepancy include optimizing domain discrepancy loss function and using adversarial discriminator. For example, joint adaptation networks (JAN) ~\cite{4} aligns the joint distributions of multiple domain-specific layers according to a joint maximum mean discrepancy (JMMD) criterion, conditional domain adversarial networks (CDANs) ~\cite{7} includes an adversarial domain discriminator conditioned on both extracted features and classifier predictions, dynamically aligning both the feature and label spaces (DAFL) ~\cite{47} calculates a dynamic weight between domain alignment and discrimination enhancement based on their conditions. However, there is no agreement on which model is the best. Learners could be complementary to achieve better performance. Thus we use the ensemble method to deal with UDA task.

\subsection{Ensemble Distillation} 
Ensemble learning uses the combination of several individual component models to obtain better generalization ability ~\cite{53, 54, 19, 20}, while it usually has large computation and memory requirements. While knowledge distillation ~\cite{55, 56, 21} is a technique to train a compact student network that it can learn the performance of a complicated teacher network. With the benefits of ensemble and distillation, ensemble distillation, which distills an ensemble model into a single student model, achieves fantastic accuracy and low computational costs. A compact student model which is trained with the information from teacher model in an ensemble learning fashion was proposed ~\cite{57, 22}. This paper extends the existing ensemble distillation methods in UDA field to adaptive nonlinear instance-aware ensemble distillation. Three techniques are used to achieve better ensemble performance in UDA, including instance-aware ensemble, shuffle linear layer, conditional domain adversarial discriminator. In addition, the theoretical proof of the validity for the proposed framework is given for better understanding.

\section{The Proposed Method}

\subsection{Overview of The Proposed IMED}

In UDA, a source domain with $n_{s}$ labeled examples is denoted as $\mathcal{D}_{s}=\left\{\left(\mathbf{x}^{i, s}, \mathbf{y}^{i, s}\right)\right\}_{i=1}^{n_{s}}$, and a target domain with $n_{t}$ unlabeled examples is denoted as $\mathcal{D}_{t}=\left\{\mathbf{x}^{j, t}\right\}_{j=1}^{n_{t}}$. It is general that the source domain and the target domain are usually sampled from different distributions, denoted as $P\left(\mathbf{x}^{s}, \mathbf{y}^{s}\right)$ and $Q\left(\mathbf{x}^{t}, \mathbf{y}^{t}\right)$ respectively, \textit{i.e.}, $P \neq Q$. There are many effective methods ~\cite{58, 59, 23, 24, 25} in UDA, such as joint adaptation networks (JAN) ~\cite{4}, conditional domain adversarial networks (CDANs) ~\cite{7}, CDANs with minimum class confusion loss and smooth domain adversarial training (CDAN+MCC+SDAT) ~\cite{15}, dynamically aligning both the feature and label spaces (DAFL) ~\cite{47}. Even with the same backbone such as ResNet-50 or ViT-B/16, different UDA models with different learning methods and variable random seeds may still present different favors of correctly classified images in the same dataset. The goal of this work is to design an ensemble method to fuse different component models' output to generate an ensembled feature, which is expected to perform better in the target domain.

\begin{figure*}[ht]
    \centering
    \includegraphics[width=180mm]{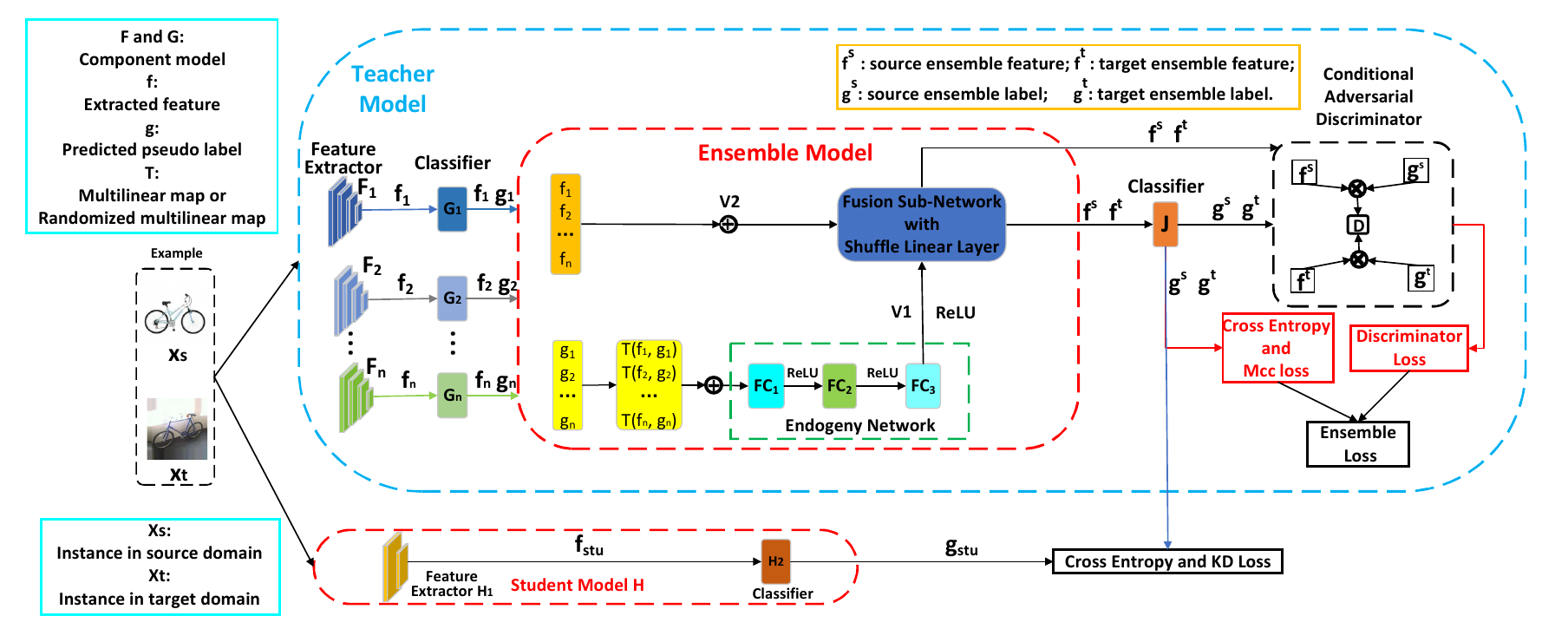}
    \caption{The architecture of Instance-aware Model Ensemble With Distillation (IMED) for unsupervised domain adaptation. $\{\mathbf{F}_{i}\}_{i=1}^{n}$ denote $n$ component models and output extracted features $\{f_{i}\}_{i=1}^{n}$ respectively. And the predicted pseudo labels $\{g_{i}\}_{i=1}^{n}$ are gained through classifier $\{\mathbf{G}_{i}\}_{i=1}^{n}$. In the ensemble model, with
    $\{f_{i}\}_{i=1}^{n}$ and $\{g_{i}\}_{i=1}^{n}$ as input, parameters of the fusion sub-network are learnt by endogeny network and the output of endogeny network is $V_{1}$. Then $\{f_{i}\}_{i=1}^{n}$ are combined in the fusion sub-network as $V_{2}$. The output of the ensemble model is the ensemble feature, which is fed to the classifier $\mathbf{J}$. The conditional adversarial discriminator will align the two domains, similar to CDANs~\cite{7}. Finally, with the well-trained ensemble teacher model, we use knowledge distillation module to get a small student model $\mathbf{H}$.}
    \label{flow}
\end{figure*}

As shown in Fig. \ref{flow}, we use $\{\mathbf{F}_{i}\}_{i=1}^{n}$ to denote  $n$ different UDA models with the same backbone and the same size of output. The difference can come from different learning methods and different random seeds. The parameters in $\{\mathbf{F}_{i}\}_{i=1}^{n}$ are denoted as $\{\theta_{\mathbf{F}_{i}}\}_{i=1}^{n}$. $f_{i}=\mathbf{F}_{i}(\mathbf{x}), (x \in \mathcal{D}_{s} \cup \mathcal{D}_{t}, 1 \leq i \leq n)$ denotes the extracted feature by $\mathbf{F}_{i}$, with the dimension $d_{f}$. Generally we use individual classifier for each component model, denoted as $\{\mathbf{G}_{i}\}_{i=1}^{n}$, with the parameters denoted as $\theta_{\mathbf{G}_{n}}$. A shared classifier sometimes is also effective for representing all the features from different component models but with less parameters, $\mathbf{G}_{i} = \mathbf{G}, (1 \leq i \leq n)$, where $\mathbf{G}$ denotes the shared classifier adopted here. Through the classifier $\{\mathbf{G}_{i}\}_{i=1}^{n}$, we can gain the predicted pseudo labels from each component model respectively, \textit{i.e.,} $g_{i} = \mathbf{G}_{i}(f_{i}), (1 \leq i \leq n)$, and the dimension is $d_{g}$. For each component model, we denote the individual loss function of each model as $\{\mathcal{L}_{i}(f_{i},g_{i})\}_{i=1}^{n}$. The training methods for the component models follow Eq. (\ref{2}).
\begin{equation}
    \begin{array}{c}
         \min\limits_{\theta_{\mathbf{F}_{1}}, \theta_{\mathbf{G}_{1}}} \mathcal{L}_{1}(f_{1},g_{1}) \\
         \min\limits_{\theta_{\mathbf{F}_{2}}, \theta_{\mathbf{G}_{2}}} \mathcal{L}_{2}(f_{2},g_{2}) \\
         \cdots\\
         \min\limits_{\theta_{\mathbf{F}_{n}}, \theta_{\mathbf{G}_{n}}} \mathcal{L}_{n}(f_{n},g_{n}) \\
    \end{array}
    \label{2}
\end{equation}

With the gained feature representations $\{f_{i}\}_{i=1}^{n}$ and predicted pseudo labels $\{g_{n}\}_{i=1}^{n}$, we can obtain the ensemble feature from the ensemble model. The details of the ensemble model will be presented in Section \ref{ensemble}, and we denote the ensemble model as $\mathbf{E}$ and its parameters as $\theta_{\mathbf{E}}$. The ensemble feature representations in the source domain and target domain are denoted as $f^{s}$ and $f^{t}$ respectively, where $f^{s} = \mathbf{E}(\{f_{i}^{s},g_{i}^{s}\}_{i=1}^{n})$ and $f^{t} = \mathbf{E}(\{f_{i}^{t},g_{i}^{t}\}_{i=1}^{n})$. The ensemble feature representations are fed to a classifier denoted by $\mathbf{J}$ which is just a fully connected linear layer, with the parameters denoted as $\theta_{\mathbf{J}}$. We can get the ensemble prediction $g^{s}$ and $g^{t}$ by $g^{s} = \mathbf{J}(f^{s})$ and $g^{t} = \mathbf{J}(f^{t})$.

Then a conditional adversarial domain discriminator conditioned on ensemble features and predictions is added to the ensemble model to reduce the domain discrepancy. The discriminator is denoted as $\mathbf{D}$ with parameters $\theta_{\mathbf{D}}$. The knowledge distillation is finally applied to generate a compact student model whose parameter size is comparable with the component model, but with better performance. The details will be shown in Section \ref{distillation}.

\subsection{Instance-aware Ensemble Model with Domain Discriminator} 
\label{ensemble}
In the ensemble model we design a neural network, named endogeny network, to learn the parameters of the fusion sub-network. The endogeny network includes $\mathbf{FC}_{1}$, $\mathbf{FC}_{2}$ and $\mathbf{FC}_{3}$. In this way, the fusion sub-network parameters are conditioned on the input instance. The details of fusion sub-network will be shown in Section \ref{sec_shuffle}.

With the feature representations $\{f_{i}\}_{i=1}^{n}$ and pseudo labels $\{g_{i}\}_{i=1}^{n}$, we firstly calculate the multilinear map between $f_{i}$ and $g_{i}$: $\mathbf{T}_{ml}(f_{i}, g_{i}) = f_{i} \otimes g_{i}$. As the dimension of $\mathbf{T}_{ml}(f_{i}, g_{i})$ is $d_{f} \times d_{g}$, which is easy to cause dimension explosion, thus we use randomized multilinear map to approximate multilinear when $d_{f} \times d_{g} \geq 4096$ ~\cite{7}. We generate random matrices $\mathbf{W}_{f} \in \mathbb{R}^{d \times d_{f}}$ and $\mathbf{W}_{g} \in \mathbb{R}^{d \times d_{g}}$, which are sampled only once and frozen when training. The elements $\mathbf{W}_{i,j}$ follow the standard normal distribution. Then the randomized multilinear map between $f_{i}$ and $g_{i}$ is $\mathbf{T}_{rml}(f_{i}, g_{i}) = \frac{1}{\sqrt{d}}(\mathbf{W}_{f}f_{i}) \odot (\mathbf{W}_{g}g_{i})$, where $\odot$ is element-wise product. Thus we have:

\begin{equation}
    \mathbf{T}(f_{i}, g_{i})= 
    \left\{
    \begin{aligned}
    \mathbf{T}_{ml}(f_{i}, g_{i}), \quad  &if d_{f} \times d_{g} \leq 4096\\ 
    \mathbf{T}_{rml}(f_{i}, g_{i}), \quad  &otherwise\\
    \end{aligned}
    \right.
\end{equation}

Then we get the input of endogeny network through the concatenation of the multilinear map as in Eq. (\ref{con1}), and the output of endogeny network acts as the parameters of fusion sub-network.

\begin{equation}
V1 = (\mathbf{T}(f_{1}, g_{1}),\mathbf{T}(f_{2}, g_{2}),\cdots,\mathbf{T}(f_{n}, g_{n}))
\label{con1}
\end{equation}

With the fusion sub-network, we can get the ensemble features $f^{s}, f^{t}$ and ensemble predictions $g^{s}, g^{t}$. Then the source domain classification loss can be calculated as Eq. (\ref{3}), and the discriminative loss is Eq. (\ref{4}). The architecture of $\mathbf{D}$ is the same as the discriminator in CDANs ~\cite{7}. It is a principled framework that conditions the adversarial adaptation models on discriminative information conveyed in the classifier predictions. It uses multilinear conditioning to capture the cross-covariance between feature representations and classifier predictions to improve the discriminability.

\begin{equation}
    \mathcal{L}_{CE}(\theta_{\mathbf{E}},\theta_{\mathbf{J}}) = \mathbb{E}_{\left(\mathbf{x}^{i, s}, \mathbf{y}^{i, s}\right) \sim \mathcal{D}_{s}}\varepsilon(g^{i, s},\mathbf{y}^{i, s})
    \label{3}
\end{equation}
\begin{equation}
    \begin{array}{cc}
    \mathcal{L}_{DC}(\theta_{\mathbf{E}}, \theta_{\mathbf{J}}, \theta_{\mathbf{D}})&=-\mathbb{E}_{\mathbf{x}^{i, s} \sim \mathcal{D}_{s}} \log \left[\mathbf{D}\left(f^{i, s}, g^{i, s}\right)\right]\\
    &-\mathbb{E}_{\mathbf{x}^{i, t} \sim \mathcal{D}_{t}} \log \left[1-\mathbf{D}\left(f^{i, t}, g^{i, t}\right)\right]
    \end{array}
    \label{4}
\end{equation}
where $\varepsilon$ is the cross-entropy loss.

Besides, it is found that reducing the class confusion leads to better performance in target domain, where the class confusion is the tendency that a classifier confuses between correct and ambiguous labels in target domain ~\cite{26}. For each batch from target domain, \textit{i.e.}, $\mathbf{X}^{t}$ denotes the input, and $\hat{\mathbf{G}}^{t} \in \mathbb{R}^{B \times |\mathcal{C}|}$ denotes the ensemble prediction through our ensemble model, where $B$ is batch size of the target data and $|\mathcal{C}|$ is the number of classes, the pairwise class confusion on the target domain is as Eq. (\ref{5}).
\begin{equation}
    \mathcal{L}_{C}(\theta_{\mathbf{E}},\theta_{\mathbf{J}}) = \mathbb{E}_{\mathbf{X}^{t} \sim \mathcal{D}_{t}}\mathcal{L}_{MCC}(\hat{\mathbf{G}^{t}})
    \label{5}
\end{equation}
where $\mathcal{L}_{MCC}$ is the minimum class confusion loss, calculated referring to ~\cite{26}.

The min-max game of conditional adversarial discriminator is as:
\begin{equation}
    \min _{\theta_{\mathbf{E}},\theta_{\mathbf{J}}} \max _{\theta_{\mathbf{D}}} \mathcal{L}_{CE}+\mu_{1} (\mathcal{L}_{C} - \mathcal{L}_{DC})
    \label{6}
\end{equation}
where $\mu_{1}$ is a hyper-parameter.

Additionally, it is found that converging to a smooth minima of a task loss (\textit{i.e.}, classification) stabilizes the adversarial training, which will lead to better generalization performance in target domain. We define the optimization objective of Smooth Domain Adversarial Training as Eq. (\ref{7}) ~\cite{15}.

\begin{equation}
    \begin{array}{cc}
    \min \limits_{\theta_{\mathbf{E}},\theta_{\mathbf{J}}} \max \limits_{\theta_{\mathbf{D}}} \max \limits_{\|\epsilon\| \leq \rho} &\mathcal{L}_{CE}(\theta_{\mathbf{E}} + \epsilon,\theta_{\mathbf{J}}+\epsilon)\\
    &+ \mu_{1} (\mathcal{L}_{C}(\theta_{\mathbf{E}} + \epsilon,\theta_{\mathbf{J}}+\epsilon)\\
    &- \mathcal{L}_{DC}(\theta_{\mathbf{E}}, \theta_{\mathbf{J}}, \theta_{\mathbf{D}})) \\
    \end{array}
    \label{7}
\end{equation}

Then we use the method called Sharpness Aware Minimization (SAM) ~\cite{27} with two gradient computation steps to derive the solution of the optimization objective Eq. (\ref{7}).

\subsection{Fusion Sub-Network with Shuffle Linear Layer}
\label{sec_shuffle}
With $V_{1}$ acting as the network parameters of fusion sub-network, we concatenate the features to generate the input $V2$ of the sub-network as:

\begin{equation}
V2 = (f_{1},f_{2},\cdots,f_{n})
\label{con2}
\end{equation}

\begin{figure}[t]
    \begin{center}
    \scalebox{1.2}{\includegraphics[width = 70mm]{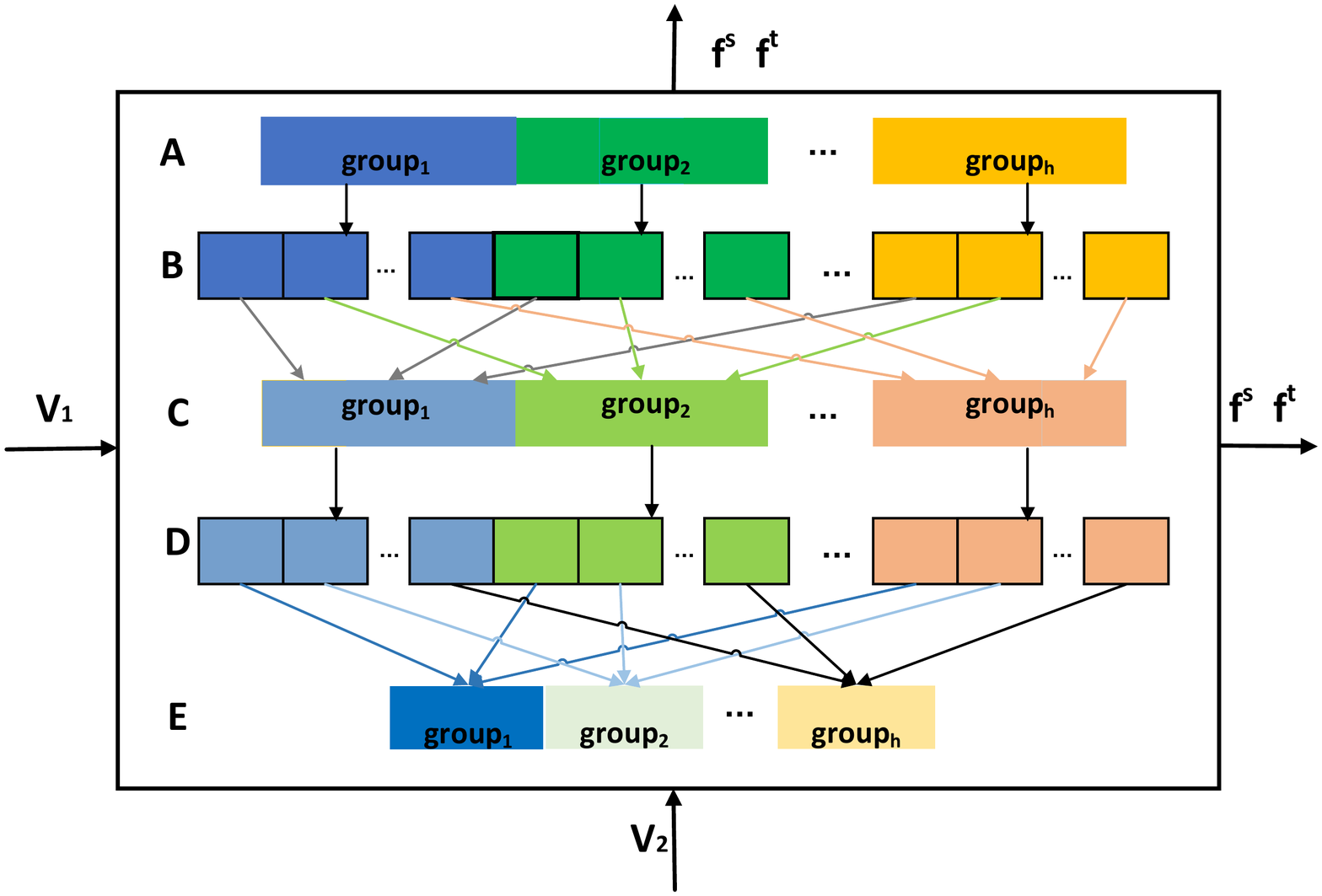}}
    \caption{Fusion Sub-Network with Shuffle Linear Layer}
    \label{shuffle}
    \end{center}
\end{figure}

In this way, the parameters of the fusion sub-network will greatly depend on the input instance. However, if the architecture of the sub-network directly uses fully connected linear layers, the number of parameters will be so large that it is hard to learn only by the endogeny network. Thus, we consider the sparse connected linear layers. As shown in Fig. \ref{shuffle}, we take position A, B, C as an example of shuffle linear layer. Firstly, we divide the input into h groups evenly (position A), and the output of the shuffle linear layer is also divided into h groups (position C). We hope that the output will utilize as much information as possible through the sparse connection, thus we divides the $group_{i} (1 \leq i \leq h)$ of input into h subgroups (as in position B), and the subgroups of $group_{i}$ will be connected to the h groups of output (position C) respectively. The relation within C, D, E is similar.

If the group number is $h$, the dimension of input is denoted as $d_{i}$ (position A), and the dimension of output is denoted as $d_{o}$ (position C). Then the overall parameters of shuffle linear layer from position A to position C are $\mathbf{P}_{num} = \frac{d_{i}d_{o}}{h}$. Thus, the number of required parameters will decrease with h. When h is small, we consider shared parameters within groups, \textit{i.e.}, the parameters of linear layer to calculate the $group_{j} (1 \leq j \leq h)$ of output is the same. In our work, we set the threshold as $\tau = 128$, \textit{i.e.,} if $h \geq \tau $, the parameters are not shared.

\begin{algorithm}[tb]
\caption{Algorithm for IMED}
\label{alg:algorithm}
\textbf{Input}: Source and target examples $\mathcal{D}_{s}, \mathcal{D}_{t}$, feature extractors $\{\mathbf{F}_{i} \}_{i=1}^{n}$ of component models, the parameters $\{\theta_{\mathbf{F}_{i}}\}_{i=1}^{n}$ are well-trained or not\\
\textbf{Parameter}: $\text{epoch}_{t}, \text{epoch}_{s}$, iters \\
\textbf{Output}: $\{\theta_{\mathbf{F}_{i}}, \theta_{\mathbf{G}_{i}}\}_{i=1}^{n}, \theta_{\mathbf{E}}, \theta_{\mathbf{J}}, \theta_{\mathbf{D}}, \theta_{\mathbf{H}_{1}}, \theta_{\mathbf{H}_{2}}$\\
\begin{algorithmic}[1] 
\STATE \% Train ensemble model as teacher model
\FOR{epoch = 1 to $\text{epoch}_{t}$}
\FOR{i = 1 to iters}
\STATE Select batch from $\mathcal{D}_{s}$ and $\mathcal{D}_{t}$
\STATE Calculate $\{\mathcal{L}_{i}\}_{i=1}^{n}$
\IF{$\{\mathbf{F_{i}}\}_{i=1}^{n}$ are well-trained at the beginning}

\STATE Update $\{\theta_{\mathbf{G}_{i}}\}_{i=1}^{n}$ respectively

\ELSE

\STATE Update $\{\theta_{\mathbf{F}_{i}}, \theta_{\mathbf{G}_{i}}\}_{i=1}^{n}$ respectively

\ENDIF

\STATE Calculate $\mathcal{L}_{CE}, \mathcal{L}_{C}, \mathcal{L}_{DC}$

\STATE Use SAM to update $\theta_{\mathbf{E}}, \theta_{\mathbf{J}}, \theta_{\mathbf{D}}$
\ENDFOR
\ENDFOR
\STATE \% Train student model
\FOR{epoch = 1 to $\text{epoch}_{s}$}
\FOR{i = 1 to iters}
\STATE Select batch from $\mathcal{D}_{s}$ and $\mathcal{D}_{t}$
\STATE Calculate $\mathcal{L}_{stu}$
\STATE Update $\theta_{\mathbf{H}_{1}}, \theta_{\mathbf{H}_{2}}$
\ENDFOR
\ENDFOR

\label{alg111}
\end{algorithmic}

\label{alg}
\end{algorithm}

\subsection{Knowledge Distillation Module \label{distill}}
\label{distillation}
When the ensemble model and domain discriminator are trained well, we can gain a teacher model and distill it into a small student model. We get ensemble feature $f^{i, s}$ and prediction $g^{i, s}$ for the source domain sample $\mathbf{x}^{i, s}$ from the teacher model, and $f^{i, t}, g^{i, t}$ for the target domain sample $\mathbf{x}^{i, t}$. The backbone $\mathbf{H_{1}}$ of student model is the same as the backbone of component models, and the classifier $\mathbf{H_{2}}$ is just a fully connected linear layer. We denote the parameters of the backbone and classifier in student model as $\theta_{\mathbf{H_{1}}}, \theta_{\mathbf{H_{2}}}$ respectively. Then we have following:

\begin{equation*}
    f_{stu}^{i, s} = \mathbf{H_{1}}(\mathbf{x}^{i, s}),
    \qquad
    f_{stu}^{i, t} = \mathbf{H_{1}}(\mathbf{x}^{i, t}),
\end{equation*}

\begin{equation*}
    g_{stu}^{i, s} = \mathbf{H_{2}}(f_{stu}^{i, s}),
    \qquad
    g_{stu}^{i, t} = \mathbf{H_{2}}(f_{stu}^{i, t}),
\end{equation*}
 
We define knowledge distillation loss (as Eq. (\ref{9})) and source classification loss (as Eq. (\ref{8})):

\begin{equation}
    \begin{array}{cc}
    \mathcal{L}_{KD}(\theta_{\mathbf{H}_{1}}, \theta_{\mathbf{H}_{2}})
    &=\mathbb{E}_{\mathbf{x}^{i, s} \sim \mathcal{D}_{s}} {\alpha}^{2} \varepsilon\left(\frac{g^{i, s}}{\alpha}, \frac{g_{stu}^{i, s}}{\alpha }  \right) \\
    &+\mathbb{E}_{\mathbf{x}^{i, t} \sim \mathcal{D}_{t}} {\alpha}^{2} \varepsilon\left(\frac{g^{i, t}}{\alpha}, \frac{g_{stu}^{i, t}}{\alpha }  \right) \\
    &+\mathbb{E}_{\mathbf{x}^{i, s} \sim \mathcal{D}_{s}} \mu_{2}  {\alpha}^{2} \varepsilon\left(\frac{f^{i, s}}{\alpha}, \frac{f_{stu}^{i, s}}{\alpha }  \right) \\
    &+\mathbb{E}_{\mathbf{x}^{i, t} \sim \mathcal{D}_{t}} \mu_{2} {\alpha}^{2} \varepsilon\left(\frac{f^{i, t}}{\alpha}, \frac{f_{stu}^{i, t}}{\alpha } \right) \\
    \end{array}
    \label{9}
\end{equation}

\begin{equation}
    \mathcal{L}_{CE_{stu}}(\theta_{\mathbf{H}_{1}}, \theta_{\mathbf{H}_{2}}) = \mathbb{E}_{\left(\mathbf{x}_{i}^{s}, \mathbf{y}_{i}^{s}\right) \sim \mathcal{D}_{s}} \varepsilon(\left(g_{stu}^{i, s}, \mathbf{y}^{i, s}\right))
    \label{8}
\end{equation}
where $\varepsilon$ is the cross-entropy loss and $\mu_{2}$ is a hyper-parameter. $\alpha$ is the distillation temperature, which is a hyper-parameter.

Then the total loss $\mathcal{L}_{stu}$ to train the student model is the summation of the above two items:

\begin{equation*}
    \mathcal{L}_{stu} = \mathcal{L}_{KD} + \mu_{3} \mathcal{L}_{CE_{stu}}
\end{equation*}
where $\mu_{3}$ is a hyper-parameter.

The overall algorithm is as Algorithm 1.

\section{Theoretical Proof For Validity}
On the theory side, the work ~\cite{28} studied how the test accuracy can be improved by ensemble or ensemble distillation of deep learning models. Here we will illustrate the validity of our framework based on their work.

\begin{lemma}{The performance of our ensemble method is better than the feature-averaging ensemble.}

\label{le1}
\end{lemma}

\emph{Proof:}
Here we take a simple case where the number of component models is 2, only one linear layer is used in shuffle linear model and the group number is 2, for derivation.

We denote the input feature representations as $f$, and $f = (f_{1},f_{2})$, where $f_{1}$ and $f_{2}$ comes from two component models respectively. According to description of Shuffle Linear Layer in the main manuscript, the output feature representations $f_{out} = W_{1}f_{1} + W_{2}f_{2}$. When all the diagonal elements in $W_{i}$ is $\frac{1}{2}$ and other elements are 0, then the feature from ensemble model is equal to the average of $f_{1}$ and $f_{2}$. Meanwhile, the elements in $W_{i} (i = 1, 2)$ are learned by the endogeny network, so the value of $W_{i}$ will be adjusted to help the ensemble model to get better performance than the case where all the diagonal elements in $W_{i}$ are $\frac{1}{2}$ and other elements are 0.
$\hfill\blacksquare$

\begin{lemma}{Superior Performance of Our Ensemble Model To Single Model.

There are n component models with the same architecture that are trained with same method and dataset. The difference is the random seed number used in initialization. Then the performance of the ensemble model of n component models in the source domain is better than any single component model.}
\label{le2}
\end{lemma}

\emph{Proof:}
When using the averaged label as an ensemble method, the performance of the ensemble model in the source domain is better than that of a single model. It has been proved ~\cite{28}.

Then we illustrate that the feature-averaging method is equal to the label-averaging method when there is only one linear layer in the classifier. We use $W$ to denote the parameters of classifier module that maps the feature to its label, the classifier is one linear layer. Here we consider a simple case with only two component models for derivation, and we denote the features extracted from two models as $f_{1}$ and $f_{2}$. And $g_{1}, g_{2}$ are labels derived from $f_{1}, f_{2}$. Thus $g_{i} = W f_{i}, i=1,2$. Then we have 
\begin{equation*}
    \frac{1}{2}(g_{1}+g_{2}) = 
    W [\frac{1}{2}(f_{1}+f_{2})]
\end{equation*}

Thus, it can be seen that the feature-averaging method in fact is the same as the label-averaging method.

Since we have shown that the performance of our proposed method is better than the feature-averaging ensemble method. And label-averaging ensemble is better than the single component model in source domain ~\cite{28}. Thus, our ensemble model can get better performance than a single component model in source domain.
$\hfill\blacksquare$



\begin{table*}
\centering
\small
\caption{Parameter setting in main experiments}
\begin{tabular}{c|c|ccccccccc}
    \hline
    Method  & Dataset &  $(\mu_{1}, \mu_{2}, \mu_{3})$ & $\alpha$  &  $h$ & $l_{0}$ & random seed & share-head-use & $\text{epoch}_{t}$ & $\text{epoch}_{s}$ & iters\\
    \hline
    {\bf $\mathbf{IMED}_{\text{CDAN+MCC+SDAT}}$} &  & (0.5, 0, 1) & 2 & 128 & 0.02 & (0, 1) & True & 20&5 &1000\\
    {\bf $\mathbf{IMED}_{\text{JAN-CDAN}}$} &  & (0.5, 1, 1) & 1 & 128 & 0.003 & (2, 2) & True & 20 & 5 & 500\\
    {\bf $\mathbf{IMED}_{\text{CDAN\_2}}$} & Office & (1, 1, 1) & 1 & 128& 0.003  & (2, 5) & True & 20 & 5 & 500\\
    {\bf $\mathbf{IMED}_{\text{CDAN\_3}}$} & -31 &(1, 0, 1)  & 1 & 128& 0.003  &(2, 5, 10) & True & 20 & 5 & 500\\
    {\bf $\mathbf{IMED}_{\text{ResNet-50}}$} &  & (0, 0, 0) & 1 & 128& 0.02  & (0, 1) & True & 20 & 5 & 500\\
    {\bf $\mathbf{IMED}_{\text{ViT}}$} &  &(0, 0, 0)  & 1 & 128& 0.02  &(0, 1) & True & 20 & 5 & 500\\
    \hline
    {\bf $\mathbf{IMED}_{\text{CDAN+MCC+SDAT}}$} &  & (1, 0, 1) & 2  & 128& 0.02 & (0, 1) & True & 15 & 5 & 1000\\
    {\bf $\mathbf{IMED}_{\text{JAN-CDAN}}$} &  & (1, 1, 1) & 1 & 128& 0.003  & (0, 0) & True& 20 & 5 & 500 \\
    {\bf $\mathbf{IMED}_{\text{CDAN\_2}}$} & Office & (1, 1, 1) & 1 & 128 & 0.003 & (0, 5) & True & 20 & 5  & 500 \\
    {\bf $\mathbf{IMED}_{\text{CDAN\_3}}$} & -home & (1, 0, 1) & 1 & 128& 0.003  & (0, 5, 10) & True & 30& 10 & 500\\
    {\bf $\mathbf{IMED}_{\text{ResNet-50}}$} &  & (0, 0, 0) & 1 & 128& 0.02  & (0, 1) & True & 20 & 5 & 500\\
    {\bf $\mathbf{IMED}_{\text{ViT}}$} &  &(0, 0, 0)  & 1 & 128& 0.02  &(0, 1) & True & 20 & 5 & 500\\
    \hline
    {\bf $\mathbf{IMED}_{\text{CDAN+MCC+SDAT}}$} &  &(0.5, 1, 1)  &2 &128 & 0.02& (0, 1) & False & 10 &5 & 500\\
    {\bf $\mathbf{IMED}_{\text{JAN-CDAN}}$} && (1, 1, 1) & 1 & 128 & 0.003 & (0, 0) & True & 20 & 5 & 500\\
    {\bf $\mathbf{IMED}_{\text{CDAN\_2}}$} &  Visda  &  (1, 1, 1) & 1 & 128& 0.003  & (0, 5) & True& 20 & 5 & 500\\
    {\bf $\mathbf{IMED}_{\text{CDAN\_3}}$} &  -2017&(1, 0, 1)  & 1 & 128 & 0.003 & (0, 5, 10) & True & 20 & 10 & 500\\
    {\bf $\mathbf{IMED}_{\text{ResNet-50}}$} &  & (0, 0, 0) & 1 & 128& 0.02  & (0, 1) & True & 20 & 5 & 500\\
    {\bf $\mathbf{IMED}_{\text{ViT}}$} &  &(0, 0, 0)  & 1 & 128& 0.02  &(0, 1) & True & 20 & 5 & 500\\
    \hline
\end{tabular}
\label{main}

\end{table*}

\begin{table*}
\centering
\small
\caption{Accuracy ($\%$) on Office-31 and Visda-2017 for unsupervised domain adaptation}
\begin{tabular}{c|cc|ccccccc|c}
    \hline
    Method  &FLOPs & Params   & A $\rightarrow$ W & D $\rightarrow$ W & W $\rightarrow$ D & A $\rightarrow$ D & D $\rightarrow$ A & W $\rightarrow$ A & Avg & Synthetic \\
     &  (G) & (M) & & & & & & & & $\rightarrow$ Real\\
    \hline
    TVT  & - & - & 96.3 & 99.4 & 100.0 & 96.4 & 85.0 & 86.0 & 93.8 & 83.9\\
    CDAN+MCC+SDAT  & 16.9 & 86.0 & \underline{97.9} & \underline{98.2} & 100.0 & \underline{84.5} & \underline{84.7} & 99.2 & \underline{94.1} & \underline{89.8}\\
    {\bf $\mathbf{IMED}_{\text{CDAN+MCC+SDAT}}$} & 16.9 & 86.0 & {\bf98.4} & {\bf 98.6} & 100.0 & {\bf 85.0} & {\bf 85.3} & 99.2 & {\bf 94.4} & {\bf 89.9}\\
    
    \hline
    JAN & 4.1 & 26.1 & 85.4 & 97.4 & 99.8 & 84.7 & 68.6 & 70.0 & \underline{84.3} & \underline{61.6}\\
    CDAN & 4.1 & 26.1 & 93.1 & 98.2 & 100.0 & 89.8 & 70.1 & 68.0 & \underline{86.6} & \underline{66.8}\\
    {\bf $\mathbf{IMED}_{\text{JAN-CDAN}}$} & 4.1  & 26.1 & 95.0 & 98.5 & 100.0 & 93.2 & 75.6 & 73.9 & {\bf 89.4}  &{\bf 74.4}\\
    {\bf $\mathbf{IMED}_{\text{CDAN\_2}}$}& 4.1 &  26.1 & 94.3 & 98.9 & 100.0 & 96.4 & 78.2 & 72.8 & {\bf 90.1}& {\bf 74.0}\\
    {\bf $\mathbf{IMED}_{\text{CDAN\_3}}$}& 4.1 & 26.1 & 93.5 & 99.0 & 99.8 & 95.0 & 76.9 & 73.5 & {\bf 89.6} & {\bf 73.2}\\
    \hline
    ResNet-50 &4.1 & 26.1 & 68.4 & 96.7 & 99.3 & 68.9 & 62.5 & 60.7 & \underline{76.1} & \underline{45.5}\\
    {\bf $\mathbf{IMED}_{\text{ResNet-50}}$} & 4.1 & 26.1 & 71.9 & 96.0 & 99.4 & 72.9 & 63.1 & 60.4 & {\bf 77.3} & {\bf 45.9}\\
    \hline
    
    ViT & 16.9 & 86.0 & 89.4 & 99.4 & 100.0 & 91.0 & 77.3 & 78.0 & \underline{89.2} & \underline{67.7}\\
    {\bf $\mathbf{IMED}_{\text{ViT}}$} &16.9 & 86.0 & 89.7 & 99.2 & 100.0 & 93.0 & 78.5 & 79.3 & {\bf 90.0} & {\bf 69.7}\\
    \hline
\end{tabular}
\label{office-31}
\end{table*}

\begin{table*}
  \small
  \centering
  \caption{Accuracy ($\%$) on Office-Home for unsupervised domain adaptation}
  \begin{tabular}{cccccccccccccc}
    \hline
    Method     & Ar  & Ar  & Ar  & Cl  & Cl  & Cl  & Pr  & Pr  & Pr  & Rw  & Rw  & Rw  &  Avg\\
    & $\rightarrow$Cl & $\rightarrow$Pr & $\rightarrow$Rw & $\rightarrow$Ar & $\rightarrow$Pr & $\rightarrow$Rw & $\rightarrow$Ar &  $\rightarrow$Cl & $\rightarrow$Rw & $\rightarrow$Ar & $\rightarrow$Cl & $\rightarrow$Pr & \\
    
    \hline
    TVT & 74.9 & 86.8 & 89.5 & 82.8 & 88.0 & 88.2 & 80.0 & 72.0 & 90.1 & 85.5 & 74.6 & 90.5 & 83.6\\
    CDAN+MCC+SDAT & \underline{70.8} & \underline{87.0} & \underline{90.5} & \underline{85.2} & \underline{87.3} & \underline{89.7} & \underline{84.1} & \underline{70.7} & \underline{90.6} & 88.3 & \underline{75.5} & \underline{92.1} & \underline{84.3}\\
    {\bf $\mathbf{IMED}_{\text{CDAN+MCC+SDAT}}$} & {\bf 72.1} & {\bf 89.0} & {\bf 90.8}& {\bf 86.2} & {\bf 88.1} & {\bf 89.8} & {\bf 84.8} & {\bf 71.7} & {\bf 91.6} & 87.8 & {\bf 76.7} & {\bf 92.4} & {\bf 85.1}\\

    \hline
    JAN & 45.9 & 61.2 & 68.9 & 50.4 & 59.7 & 61.0 & 45.8 & 43.4 & 70.3 & 63.9 & 52.4 & 76.8 & \underline{58.3}\\
    CDAN & 49.0 & 69.3 & 74.5 & 54.4 & 66.0 & 68.4 & 55.6 & 48.3 & 75.9 & 68.4 & 55.4 & 80.5 & \underline{63.8}\\
    {\bf $\mathbf{IMED}_{\text{JAN-CDAN}}$}& 53.3 & 72.9 & 77.8 & 61.4 & 70.5 & 71.4 & 61.3 & 53.1 & 78.7 & 74.0 & 59.9 & 83.2 & {\bf 68.1}\\
    {\bf $\mathbf{IMED}_{\text{CDAN\_2}}$} & 55.6 & 72.1 & 78.4 & 61.7 & 71.8 & 72.3 & 64.6 &55.5 & 79.8 & 74.2 & 60.1 & 82.9 & {\bf 69.1}\\
    {\bf $\mathbf{IMED}_{\text{CDAN\_3}}$} &  56.8 & 73.9 & 78.7 & 62.2 & 72.1 & 72.9 & 64.9 & 54.8 & 80.1 & 75.4 & 61.5 & 83.9 & {\bf 69.8} \\
    \hline
    ResNet-50 & 34.9 & 50.0 & 58.0& 37.4 & 41.9 & 46.2 & 38.5 & 31.2 & 60.4 & 53.9 & 41.2 & 59.9 & \underline{46.1}\\
    {\bf $\mathbf{IMED}_{\text{ResNet-50}}$} & 40.0 & 62.1 & 66.1& 47.2 & 53.8 & 57.3 & 45.4 & 38.0 & 65.0 & 60.9 & 42.9 & 74.2 & {\bf 54.4}\\
    \hline
    ViT & 52.4 & 82.1 & 86.9 & 76.8 & 84.1 & 86.0 & 75.1 & 51.2 & 88.1 & 78.3 & 51.5 & 87.8 & \underline{75.0}\\
    {\bf $\mathbf{IMED}_{\text{ViT}}$} & 52.9 & 80.8 & 85.3 & 76.6 & 82.4 & 85.6 & 75.9 & 54.1 & 87.1 & 79.1 & 53.8 & 88.0 & {\bf 75.1}\\
    \hline
  \end{tabular}
  \label{office-home}
\end{table*}





\begin{table*}[t]
\small
\centering
\caption{Ablation Study: Accuracy ($\%$) on Office-31 (ResNet-50)}
\begin{tabular}{c|cc|ccccccc}
    \hline
    Method    & FLOPs (G) & Params (M) & A $\rightarrow$ W & D $\rightarrow$ W & W $\rightarrow$ D & A $\rightarrow$ D & D $\rightarrow$ A & W $\rightarrow$ A & Avg\\
    \hline
    JAN  & - & -& 85.4 & 97.4 & 99.8 & 84.7 & 68.6 & 70.0 & \underline{84.3} \\
    CDAN & - &- & 93.1 & 98.2 & 100.0 & 89.8 & 70.1 & 68.0 & \underline{86.6} \\
    {\bf $\mathbf{IMED}_{\text{JAN-CDAN}}$} & {\bf 4.1} & {\bf 26.1} & 95.0 & 98.5 & 100.0 & 93.2 & 75.6 & 73.9 & {\bf 89.4} \\
    \hline
    
    {\bf $\mathbf{IMED}_{\text{non\_aware}}$} & - & - & 94.5 & 98.5 & 98.9 & 91.4 & 75.7 & 74.4 & 88.9 \\
    \hline
    
    {\bf $\mathbf{IMED}_{\text{avg}}$} & - & - & 93.8 & 98.2 & 100.0 & 92.8 & 73.9 & 74.4 & 88.8 \\
    \hline
    
    {\bf $\mathbf{IMED}_{\text{FCnet}}$} & - & - & 94.7 & 98.5 &100.0 & 91.4 & 73.3 & 73.2 & 88.5 \\
    \hline
    
    {\bf $\mathbf{IMED}_{\text{o\_distill}}$} & \underline{8.2} & \underline{52.2} & 94.3 & 98.2 & 100.0 & 93.2 & 74.7 & 74.0 & 89.0 \\
    \hline

    {\bf $\mathbf{IMED}_{\text{h\_128}}$} & - & - & 95.0 & 98.5 & 100.0 & 93.2 & 75.6 & 73.9 & {\bf 89.4}   \\
    {\bf $\mathbf{IMED}_{\text{h\_32}}$} & - & - & 93.8 & 98.1 & 100.0 & 92.4 & 73.2 & 73.8 & 88.6 \\
    {\bf $\mathbf{IMED}_{\text{h\_16}}$} & - & - & 93.8 & 98.6 & 100.0 & 93.4 & 73.1 & 73.2 & 88.7 \\
    \hline
    {\bf $\mathbf{IMED}_{\text{shuffle\_1}}$} & - & - & 94.8 & 98.9 & 100.0 & 93.6 & 74.4 & 72.5 & 89.0  \\
    {\bf $\mathbf{IMED}_{\text{shuffle\_2}}$} & - & - & 95.0 & 98.5 & 100.0 & 93.2 & 75.6 & 73.9 & {\bf 89.4} \\
    {\bf $\mathbf{IMED}_{\text{shuffle\_3}}$} & - & - & 95.8 & 98.5 & 100.0 & 93.6 & 74.4 & 72.5 &  89.1\\
    \hline
\end{tabular}
\label{ablation1}
\end{table*}

\begin{table*}
  \small
  \centering
  \caption{Ablation Study: Accuracy ($\%$) on Office-Home (ResNet-50)}
  \begin{tabular}{cccccccccccccc}
    \hline
    Method    & Ar  & Ar  & Ar  & Cl  & Cl  & Cl  & Pr  & Pr  & Pr  & Rw  & Rw  & Rw  &  Avg\\
    & $\rightarrow$Cl & $\rightarrow$Pr & $\rightarrow$Rw & $\rightarrow$Ar & $\rightarrow$Pr & $\rightarrow$Rw & $\rightarrow$Ar &  $\rightarrow$Cl & $\rightarrow$Rw & $\rightarrow$Ar & $\rightarrow$Cl & $\rightarrow$Pr & \\
    
    \hline
    
    JAN & 45.9 & 61.2 & 68.9 & 50.4 & 59.7 & 61.0 & 45.8 & 43.4 & 70.3 & 63.9 & 52.4 & 76.8 & \underline{58.3}\\
    CDAN  & 49.0 & 69.3 & 74.5 & 54.4 & 66.0 & 68.4 & 55.6 & 48.3 & 75.9 & 68.4 & 55.4 & 80.5 & \underline{63.8}\\
    {\bf $\mathbf{IMED}_{\text{JAN-CDAN}}$} & 53.3 & 72.9 & 77.8 & 61.4 & 70.5 & 71.4 & 61.3 & 53.1 & 78.7 & 74.0 & 59.9 & 83.2 & {\bf 68.1}\\
    \hline
    
    {\bf $\mathbf{IMED}_{\text{non\_aware}}$}  &  52.4 & 72.4 & 76.8 & 60.2 & 70.4 & 69.9 & 60.7 & 51.4 &77.5 & 71.9 & 57.9 & 81.5 & 66.9\\
    \hline
    
    {\bf $\mathbf{IMED}_{\text{avg}}$}  & 52.4 & 71.1 & 76.9 & 60.5 & 69.0 & 67.6 & 61.4 & 52.7 & 78.2 & 74.2 & 59.0 & 83.2 & 67.1 \\
    \hline
    
    {\bf $\mathbf{IMED}_{\text{FCnet}}$}  & 53.2 & 72.6 & 77.0 & 61.1 & 70.1 & 70.0 & 61.6 & 53.0 & 78.4 & 73.0 & 57.9 & 83.2 & 67.6 \\
    \hline
    
    {\bf $\mathbf{IMED}_{\text{o\_distill}}$} &  53.1 & 72.6 & 77.8 & 61.3 & 70.2 & 71.3 & 61.0 & 53.0 & 78.4 & 73.7 & 59.8 & 83.0 & 67.9\\
    \hline

    {\bf $\mathbf{IMED}_{\text{h\_128}}$}  & 53.3 & 72.9 & 77.8 & 61.4 & 70.5 & 71.4 & 61.3 & 53.1 & 78.7 & 74.0 & 59.9 & 83.2 & {\bf 68.1} \\
    {\bf $\mathbf{IMED}_{\text{h\_32}}$}  & 51.8 & 73.1 & 77.4 & 60.5 & 71.1 & 69.8 & 61.5 & 53.3 & 77.4 & 74.0 & 59.0 & 83.3 & 67.7 \\
    {\bf $\mathbf{IMED}_{\text{h\_16}}$}  & 55.1 & 73.2 & 78.1 & 60.8 & 70.4 & 71.4 & 60.7 & 52.3 & 78.1 & 73.8 & 58.7 & 82.0& 67.8 \\
    \hline
    {\bf $\mathbf{IMED}_{\text{shuffle\_1}}$}  & 54.1 & 73.0 & 78.2 & 60.7 & 71.1 & 71.2 & 62.4 & 53.7 & 78.1 & 73.4 & 58.9 & 81.4 & 68.0  \\
    {\bf $\mathbf{IMED}_{\text{shuffle\_2}}$}  & 53.3 & 72.9 & 77.8 & 61.4 & 70.5 & 71.4 & 61.3 & 53.1 & 78.7 & 74.0 & 59.9 & 83.2 & {\bf 68.1}   \\
    
    {\bf $\mathbf{IMED}_{\text{shuffle\_3}}$}  & 54.0 & 73.1 & 77.9 & 60.7 & 71.2 & 71.3 & 61.5 & 53.3 & 78.4 & 72.9 & 59.8 & 82.0 & 68.0\\
    \hline
  \end{tabular}
  \label{ablation2}
\end{table*}

\begin{figure*}[ht]
\centering
\includegraphics[width=180mm]{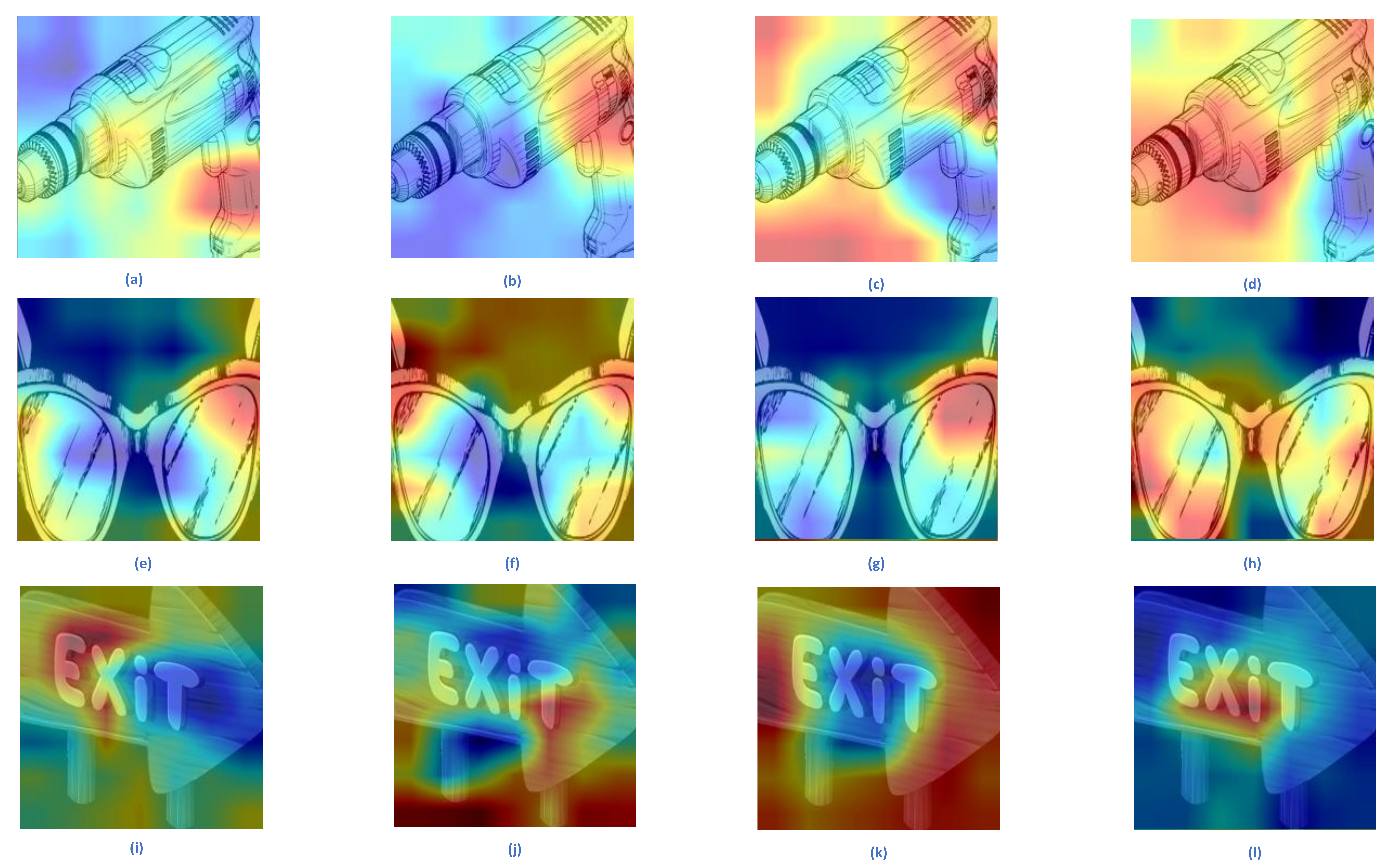}
\caption{Feature map visualization: In each row, four figures (from left to right) respectively represents CDANs, JAN, averaging ensemble distillation model, our proposed ensemble distillation model. And the category of three rows is drill, glasses and exit sign respectively.}
\label{vis}
\end{figure*}

\begin{figure*}[ht]
\centering
\includegraphics[width=180mm]{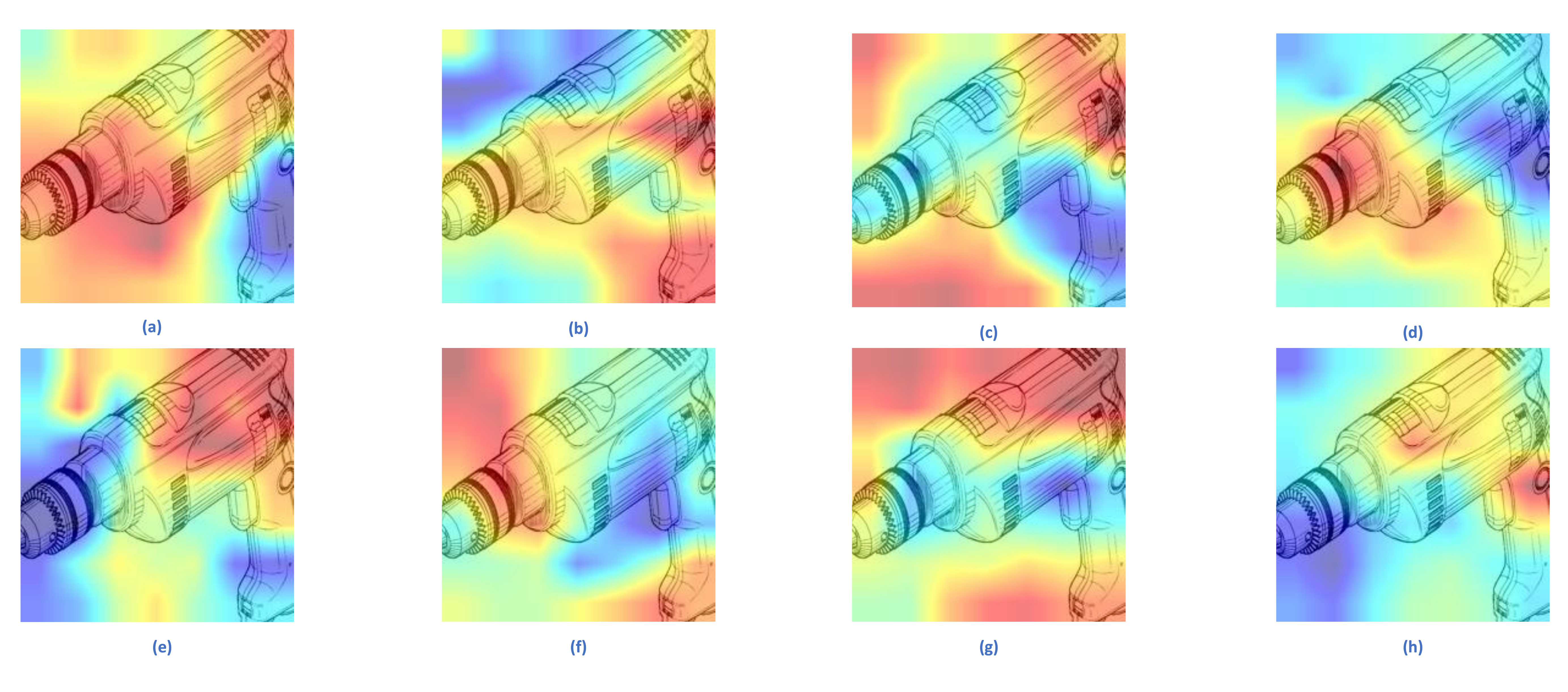}
\caption{Feature maps of drill from (a) $\mathbf{IMED}_{\text{JAN-CDAN}}$, 
(b) $\mathbf{IMED}_{\text{non\_aware}}$, (c) $\mathbf{IMED}_{\text{avg}}$, 
(d) $\mathbf{IMED}_{\text{FCnet}}$, (e) $\mathbf{IMED}_{\text{h\_32}}$, 
(f) $\mathbf{IMED}_{\text{h\_16}}$, (g) $\mathbf{IMED}_{\text{shuffle\_1}}$, 
(h) $\mathbf{IMED}_{\text{shuffle\_3}}$, respectively. And (a) represents model $\mathbf{IMED}_{\text{h\_128}}$ or $\mathbf{IMED}_{\text{shuffle\_2}}$ as well.}
\label{vis_class1}
\end{figure*}

\begin{figure*}[ht]
\centering
\includegraphics[width=180mm]{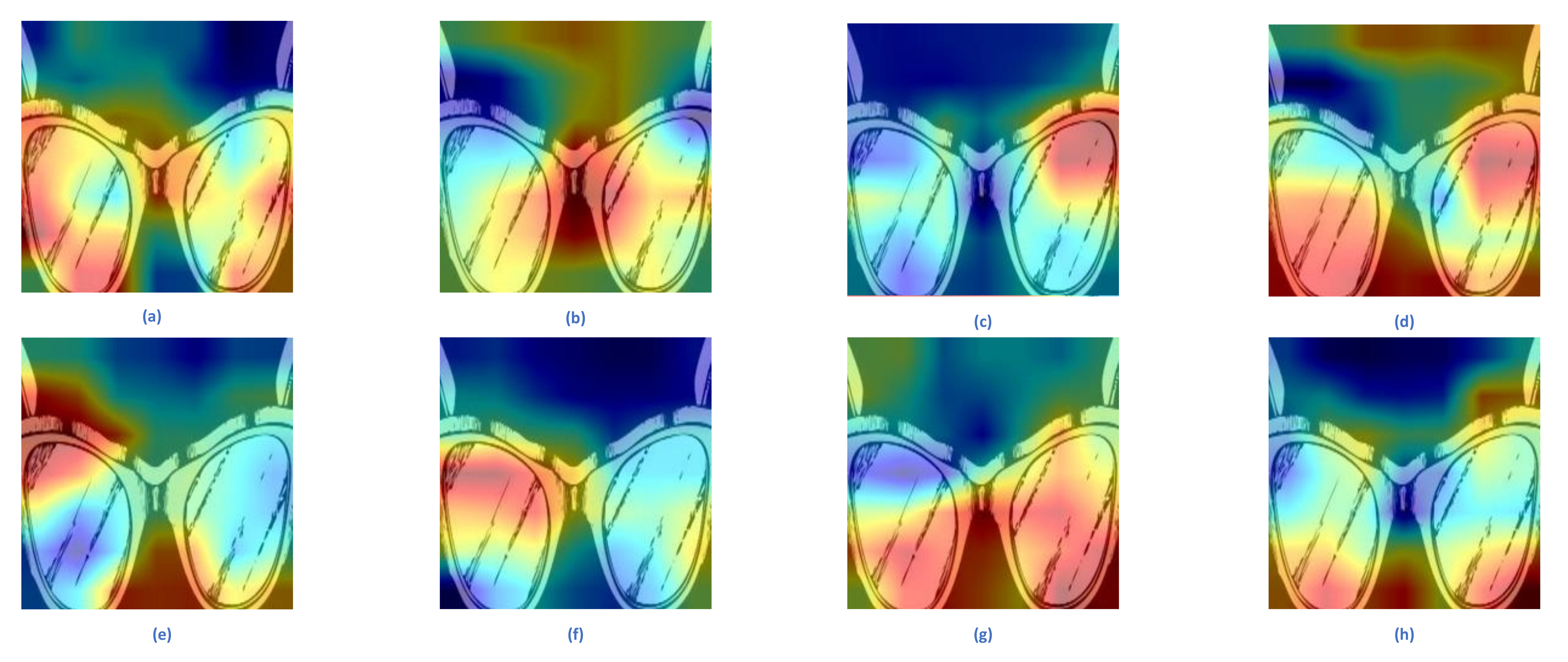}
\caption{Feature maps of glasses from (a) $\mathbf{IMED}_{\text{JAN-CDAN}}$, 
(b) $\mathbf{IMED}_{\text{non\_aware}}$, (c) $\mathbf{IMED}_{\text{avg}}$, 
(d) $\mathbf{IMED}_{\text{FCnet}}$, (e) $\mathbf{IMED}_{\text{h\_32}}$, 
(f) $\mathbf{IMED}_{\text{h\_16}}$, (g) $\mathbf{IMED}_{\text{shuffle\_1}}$, 
(h) $\mathbf{IMED}_{\text{shuffle\_3}}$, respectively. And (a) represents model $\mathbf{IMED}_{\text{h\_128}}$ or $\mathbf{IMED}_{\text{shuffle\_2}}$ as well.}

\label{vis_class2}
\end{figure*}

\begin{figure*}[ht]
\centering
\includegraphics[width=180mm]{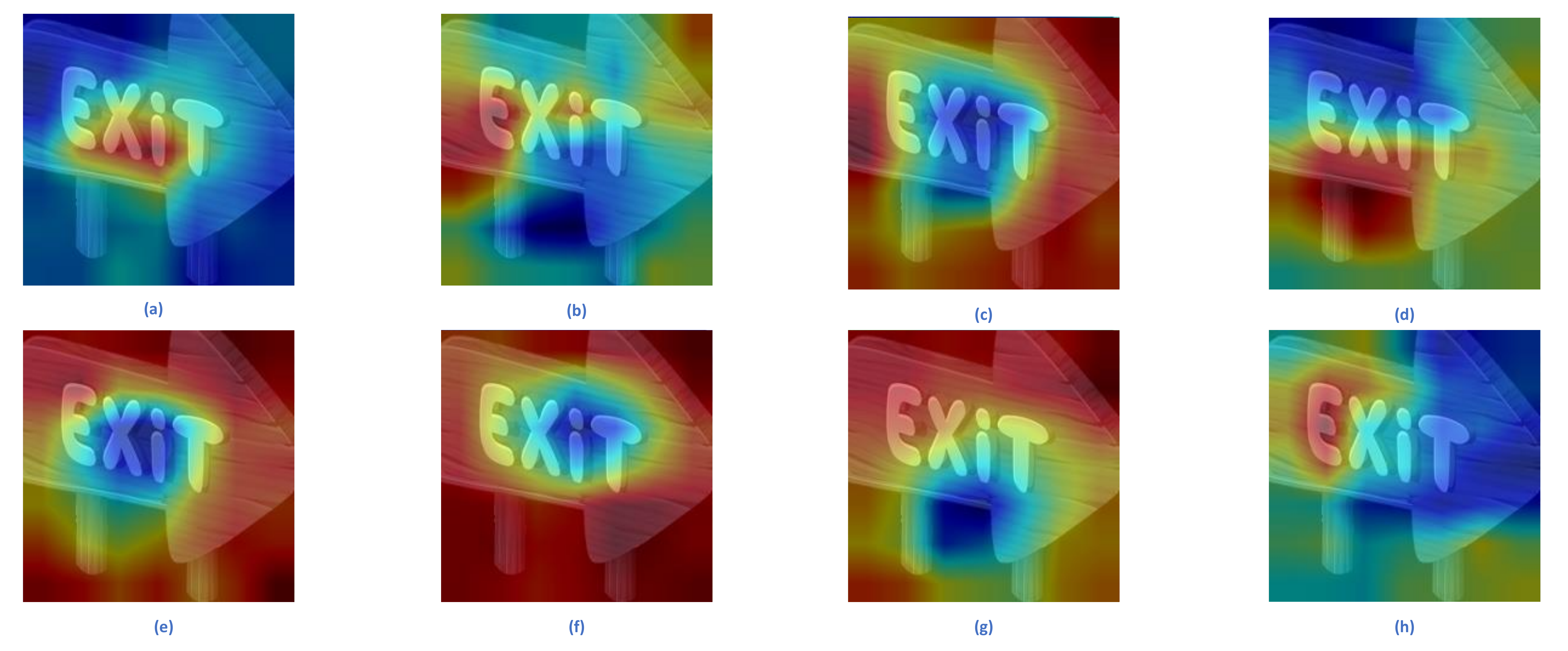}
\caption{Feature maps of exit sign from (a) $\mathbf{IMED}_{\text{JAN-CDAN}}$, 
(b) $\mathbf{IMED}_{\text{non\_aware}}$, (c) $\mathbf{IMED}_{\text{avg}}$, 
(d) $\mathbf{IMED}_{\text{FCnet}}$, (e) $\mathbf{IMED}_{\text{h\_32}}$, 
(f) $\mathbf{IMED}_{\text{h\_16}}$, (g) $\mathbf{IMED}_{\text{shuffle\_1}}$, 
(h) $\mathbf{IMED}_{\text{shuffle\_3}}$, respectively. And (a) represents model $\mathbf{IMED}_{\text{h\_128}}$ or $\mathbf{IMED}_{\text{shuffle\_2}}$ as well.}
\label{vis_class3}
\end{figure*}

\begin{table*}
\centering
\small
\caption{Ablation Experiment To Check Loss Adjustment Factors: Accuracy ($\%$) on Office-31 (ResNet-50)}
\begin{tabular}{c|ccccccc}
    \hline
    $(\mu_{1}, \mu_{2}, \mu_{2})$ & A $\rightarrow$ W & D $\rightarrow$ W & W $\rightarrow$ D & A $\rightarrow$ D & D $\rightarrow$ A & W $\rightarrow$ A & Avg\\
    \hline
    $(1, 1, 1)$  & 94.3 & 98.6 & 100.0 & 93.4 & 74.8 & 74.1 & \underline{89.2}  \\
    $(0.5, 1, 1)$  & 95.0 & 98.5 & 100.0 & 93.2 & 75.6 & 73.9 & {\bf 89.4} \\
    $(1, 0.5, 1)$  & 95.0  & 98.6 & 100.0 & 93.2 & 75.4 & 73.8 &  {\bf 89.3}\\
    $(1, 1, 0.5)$  &  93.1 & 98.7  & 100.0 & 94.2 & 77.0 & 71.2 & 89.0  \\
    $(3, 1, 1)$  & 94.8  &  98.5 & 100.0 & 93.6 & 75.4 & 73.9 & 89.1 \\
    $(1, 3, 1)$  & 93.3 & 98.6 & 100.0 & 93.2 & 75.4 & 73.8 & 89.0 \\
    $(1, 1, 3)$  & 93.2 & 98.9  & 100.0 &  94.6 & 77.3 & 72.7 & {\bf 89.4} \\
    \hline
\end{tabular}
\label{factor}
\end{table*}

\begin{table*}
\centering
\small
\caption{Ablation Experiment To Check Distillation Temperature: Accuracy ($\%$) on Office-31 (ResNet-50)}
\begin{tabular}{c|ccccccc}
    \hline
    $\alpha$ & A $\rightarrow$ W & D $\rightarrow$ W & W $\rightarrow$ D & A $\rightarrow$ D & D $\rightarrow$ A & W $\rightarrow$ A & Avg\\
    \hline
    1  & 95.0 & 98.5 & 100.0 & 93.2 & 75.6 & 73.9 & {\bf 89.4} \\
    2  &  93.7 & 98.5 & 100.0 & 92.4  & 73.6 & 73.8 & 88.7\\ 
    3  & 93.5 & 98.5 & 100.0  & 92.4 & 73.5 & 73.7 & 88.6\\ 
    \hline
\end{tabular}
\label{distill1}
\end{table*}

\begin{theorem}The Validity Of Our Framework

We define the source risk as 
\begin{equation*}
    R_{S}=\mathbb{E}_{\left(\mathbf{x}^{i, s}, \mathbf{y}^{i, s}\right) \sim \mathcal{D}_{s}}\left[g_{stu}^{i, s} \neq \mathbf{y}^{i, s}\right]
\end{equation*}

and the target risk as 

\begin{equation*}
    R_{T}=\mathbb{E}_{\left(\mathbf{x}^{i, t}, \mathbf{y}^{i, t}\right) \sim \mathcal{D}_{t}}\left[g_{stu}^{i, t} \neq \mathbf{y}^{i, t}\right]
\end{equation*}

Then our proposed ensemble distillation model will help to decrease the target risk.

\end{theorem}

\emph{Proof}: 
We formalize the problem of domain adaptation for binary classification. Let $\mathcal{H}$ be a hypothesis space of $V C$ dimension d, $\mathcal{D}_{s}, \mathcal{D}_{t}$ denote samples of size $m^{\prime}$ drawn from $P$ and $Q$ respectively.

We define the source risk of $h \in \mathcal{H}$ as 
\begin{equation*}
    R_{S}(h)=\mathbb{E}_{\left(\mathbf{x}^{i, s}, \mathbf{y}^{i, s}\right) \sim P}\left[h(\mathbf{x}^{i, s}) \neq \mathbf{y}^{i, s}\right]
\end{equation*}

and the target risk of $h$ as 

\begin{equation*}
    R_{T}(h)=\mathbb{E}_{\left(\mathbf{x}^{i, t}, \mathbf{y}^{i, t}\right) \sim Q}\left[h(\mathbf{x}^{i, t}) \neq \mathbf{y}^{i, t}\right].
\end{equation*}

It has been shown that $R_{T}(h)$ can be controled by $R_{S}(h)$ and the $\mathcal{H}\Delta\mathcal{H}$-divergence between $\mathcal{D}_{s}, \mathcal{D}_{t}$ ~\cite{32}. From \emph{Lemma 2}, we can see that the ensemble model in our framework can decrease the source risk $R_{S}(h)$, and the conditional adversarial discriminator will help to decrease the $\mathcal{H}\Delta\mathcal{H}$-divergence between $\mathcal{D}_{s}, \mathcal{D}_{t}$. Thus our proposed ensemble distillation model will help to decrease the target risk.
$\hfill\blacksquare$

\section{Experiments}
We select following three datasets for evaluation: \textbf{Office-31} ~\cite{32}, \textbf{Office-Home} ~\cite{33}, and \textbf{VisDA-2017} \footnote{\url{http://ai.bu.edu/visda-2017/}}. We evaluate the proposed method based on six ensemble baselines as follows, and compare the finally distilled student model from these ensemble models with the original component models respectively. The FLOPs, Params and classification accuracy are presented to show the effectiveness of the proposed method. 
\begin{itemize}
    \item$\mathbf{IMED}_{\text{CDAN+MCC+SDAT}}$: We use two Conditional Domain Adversarial Networks with minimum class confusion and smooth domain adversarial training (CDAN+MCC+SDAT) ~\cite{15} under different initialization random numbers as the component models. The backbone of CDAN+MCC+SDAT and student model is \textbf{ViT-B/16} ~\cite{41}.
    \item$\mathbf{IMED}_{\text{JAN-CDAN}}$: We use Joint Adaptation Network (JAN) ~\cite{4} and Conditional Domain Adversarial Networks (CDANs) ~\cite{7} with the same initialization random number as the component models. The backbone of JAN, CDANs and student model is \textbf{ResNet-50}.
    \item$\mathbf{IMED}_{\text{CDAN\_2}}$: We use two CDANs with different initialization random numbers as the component models, and the backbone network of CDANs and student model is \textbf{ResNet-50}.
    \item$\mathbf{IMED}_{\text{CDAN\_3}}$: We use three CDANs with different initialization random numbers as the component models, and the backbone network of CDANs and student model is \textbf{ResNet-50}.
    \item$\mathbf{IMED}_{\text{ResNet-50}}$: We use two ResNet-50 with different initialization random numbers as the component models, to validate our method on source-only model \textbf{ResNet-50}. For the source-only model, we only have the information about source domain $\mathcal{D}_{s}$. So we do not calculate the loss function term $\mathcal{L}_{C}, \mathcal{L}_{DC}$, the term of $\mathcal{L}_{stu}$ associated with $\mathcal{D}_{t}$ in Algorithm \ref{alg:algorithm}, other framework is the same as Algorithm \ref{alg:algorithm}.
    \item$\mathbf{IMED}_{\text{ViT}}$: We use two ViT-B/16 with different initialization random numbers as the component models, to validate our method on source-only model \textbf{ViT-B/16}.
\end{itemize}

All the backbones are initialised with ImageNet pretrained weights. And we choose PyTorch ~\cite{60} with version 1.10.0 as the framework to implement all algorithms. All the experimental results are obtained with RTX3090 (NVIDIA GeForce RTX 3090) GPU clusters.

\subsection{Detailed Setting In Experiments}

For all the experiments, we use batch size as 32. And we adopt SGD with 0.9 as the momentum, and we use learning rate annealing strategy as ~\cite{8}: the learning rate is adjusted as $l = l_{0}(1 + \hat{\beta}p)^{-\beta}$, where $l_{0} = 0.01, \hat{\beta} = 10, \beta = 0.75$, and p is the training process changing from 0 to 1. 

The parameters used in experiments are as follows:
\begin{itemize}
    \item $(\mu_{1}, \mu_{2}, \mu_{3})$: the loss adjustment factors among the loss items;
    \item $\alpha$: the temperature in knowledge distillation;
    \item $h$: the group number in shuffle linear layer;
    \item $l_{0}$: initial value for the learning rate annealing strategy;
    \item random seed: the initial random seeds for the component models, the format is $(\text{seed}_{1}, \text{seed}_{2}, \dots, \text{seed}_{n})$, where $\text{seed}_{i}$ denotes the $i$-th component model's initial random seed;
    \item share-head-use: whether to use the shared head for all the component models, True denotes that the heads are shared;
    \item $\text{epoch}_{t}$: the total number of epochs when training the ensemble teacher model;
    \item $\text{epoch}_{s}$: the total number of epochs when training the student model;
    \item iters: the total number of iterations in each epoch when training ensemble teacher model or student model;
\end{itemize}

The specific parameter setting in main experiments is shown in TABLE \ref{main}.

\subsection{Results}
The results on Office-31 and Visda-2017 are presented in TABLE \ref{office-31}. Three models including ViT ~\cite{35}, TVT ~\cite{36} and CDAN+MCC+SDAT ~\cite{15} with the same backbone ViT-B/16 are selected for comparison. It can be found that with comparable computational cost, $\mathbf{IMED}_{\text{CDAN+MCC+SDAT}}$ exceeds the state-of-the-art methods like CDAN+MCC+SDAT, and achieves 94.4\% on Office-31 and 89.9\% on Visda-2017. Additionally, the ResNet-50, JAN and CDAN with the same ResNet-50 backbone are also compared with our method. It can be seen that compared with the original component models, the student models distilled by our method achieve consistently better performance, \textit{i.e.}, $\mathbf{IMED}_{\text{CDAN+MCC+SDAT}}$, $\mathbf{IMED}_{\text{JAN-CDAN}}$, $\mathbf{IMED}_{\text{CDAN\_2}}$, $\mathbf{IMED}_{\text{CDAN\_3}}$, $\mathbf{IMED}_{\text{ResNet-50}}$, and $\mathbf{IMED}_{\text{ViT}}$ increase the accuracy by $0.3\%, 2.8\%, 3.5\%, 3.0\%, 1.2\%$, and $0.8\%$ respectively on Office-31. On Visda-2017, they increase the accuracy by $0.1\%, 7.6\%, 7.2\%, 6.4\%, 0.4\%$, and $2.0\%$ respectively.

The results on Office-Home are reported in TABLE \ref{office-home}. We show that with similar computational complexity (refer to TABLE \ref{office-31} for FLOPs and Params), $\mathbf{IMED}_{\text{CDAN+MCC+SDAT}}$ exceeds the state-of-the-art methods like CDAN+MCC+SDAT, and achieves 85.1\% on Office-home. Compared with the original component models, student models distilled by our method achieve consistently better performance, \textit{i.e.}, $\mathbf{IMED}_{\text{CDAN+MCC+SDAT}}$, $\mathbf{IMED}_{\text{JAN-CDAN}}$, $\mathbf{IMED}_{\text{CDAN\_2}}$, $\mathbf{IMED}_{\text{CDAN\_3}}$, $\mathbf{IMED}_{\text{ResNet-50}}$, and $\mathbf{IMED}_{\text{ViT}}$ increase the accuracy by $0.8\%, 4.3\%, 5.3\%$, $6.0\%$, $8.3\%$, and $0.1\%$ respectively.



\subsection{Ablation Study}
The ablation experiments are performed on the Office-31 and Office-Home with model $\mathbf{IMED}_{\text{JAN-CDAN}}$, the results are shown in TABLE \ref{ablation1} and TABLE \ref{ablation2} respectively. In particular, the ablative settings include following groups: 

\textbf{Instance-aware vs. Non-instance-aware:}
We compare the performance of instance-aware model versus non-instance-aware model, where the parameters of the fusion sub-network are trained to be identical for all the instances and denoted as $\mathbf{IMED}_{\text{non\_aware}}$. Results show that instance-aware strategy brings $0.5\%$ (Office-31) and $1.2\%$ (Office-Home) accuracy increase as compared with non-instance-aware one.

\textbf{Non-linear vs. Averaging:}
We compare the performance of the non-linear ensemble model with the averaging method where the mean of component models' features is computed as the ensemble feature, which is denoted as $\mathbf{IMED}_{\text{avg}}$. Results show that the non-linear strategy brings $0.6\%$ (Office-31) and $1.0\%$ (Office-Home) accuracy increase over the averaging one.

\textbf{Shuffle linear vs. Fully connected linear:} We use a typical three-layer fully connected network to replace the shuffle linear module, denoted as $\mathbf{IMED}_{\text{FCnet}}$. Results show that the shuffle linear brings $0.9\%$ (Office-31) and $0.5\%$ (Office-Home) accuracy increase.

\textbf{Distillation vs. Non-distillation:} We report the FLOPs (TABLE \ref{ablation1}), Params (TABLE \ref{ablation1}), and classification accuracy of the ensemble model without distillation, which is denoted as $\mathbf{IMED}_{\text{o\_distill}}$. Results show that distillation-based model brings $0.4\%$ (Office-31), $0.2\%$ (Office-Home) accuracy increase and decreases $50\%$ FLOPs and Params.

\textbf{Group number in shuffle linear layer:} We examine the influence of the  group number in shuffle linear layer by setting its value $h = 128, 32, 16$. The model is denoted as $\mathbf{IMED}_{\text{h\_128}}, \mathbf{IMED}_{\text{h\_32}}, \mathbf{IMED}_{\text{h\_16}}$ respectively. The performance reaches the best when $h = 128$.

\textbf{Depth of fusion sub-network:} We examine the performance of the fusion sub-network with one, two or three shuffle linear layers, denoted as 
$\mathbf{IMED}_{\text{shuffle\_1}}$, $\mathbf{IMED}_{\text{shuffle\_2}}$, $\mathbf{IMED}_{\text{shuffle\_3}}$. Results how that performance is the best when there are two shuffle linear layers.

\subsection{Feature Map Visualization}
We visualize the feature map to show the validity of our proposed ensemble distillation model IMED, and the better performance compared to other models in ablation study. Here we use the model with ResNet-50 as backbone, and perform domain adaptation from Product images ({\bf Pr}) to Clip Art ({\bf Cl}) of dataset Office-Home.

\textbf{Validity Of IMED}
We visualize the feature map from model CDANs, JAN, averaging ensemble distillation model of CDANs and JAN, and our proposed ensemble distillation model of CDANs and JAN. The results can be seen in Fig. \ref{vis}. Compared with CDANs, JAN or averaging ensemble distillation model, our proposed ensemble model can capture more important object features and focus less on the background environment. Thus our proposed method IMED can get better performance.

\textbf{Compared With Other Models In Ablation Study}
In Fig. \ref{vis_class1}, Fig. \ref{vis_class2}, Fig. \ref{vis_class3}, we can find that $\mathbf{IMED}_{\text{JAN-CDAN}}$ can capture more features of the object and suppress more areas of the background than other models.

\subsection{Ablation Experiments For Loss Adjustment Factors}

In the experiment, we want to check the influence of different loss adjustment factors $(\mu_{1}, \mu_{2}, \mu_{3})$. We use the ensemble distillation model of CDANs and JAN, $\mathbf{IMED}_{\text{JAN-CDAN}}$, to do experiments on dataset Office-31 with ResNet-50 as the backbone. The results are shown in TABLE \ref{factor}. We can find that the decrease of the proportion for $(\mu_{1}, \mu_{2})$ or the increase of the proportion for $\mu_{3}$ has some positive but limited boost to the performance of ensemble distillation model.

\subsection{Ablation Experiments For Distillation Temperature}

In the experiment, we want to check the influence of distillation temperature $\alpha$. We use the ensemble distillation model of CDANs and JAN, $\mathbf{IMED}_{\text{JAN-CDAN}}$, to do experiments on dataset Office-31 with ResNet-50 as the backbone. The results are shown in TABLE \ref{distill1}. We can find that the increase of the temperature will bring a lower accuracy of the final student model.

\section{Conclusion}
The paper presents an Instance-aware Model Ensemble With Distillation (IMED) framework for UDA. Unlike previous works that adopt simple linear ensemble-based strategies, IMED fuses diversified UDA-based component models via an instance-aware non-linear strategy. Three key components: {\bf Instance-aware module} (to utilize instance-level representation characteristics in ensemble model), {\bf Non-linear module} (to create a diversified and large enough ensemble feature space), {\bf Shuffle linear layer} (to decrease the parameters required in fusion sub-network) make the main contributions to improve the performance. And {\bf Knowledge distillation module} contributes to reduce the computational cost when deploying model. Experiments based on several ensemble baselines and UDA datasets validate the effectiveness of the proposed framework.

\bibliographystyle{IEEEtran}
\bibliography{main}

\vfill

\end{document}